# Predictive Coding-based Deep Dynamic Neural Network for Visuomotor Learning

Jungsik Hwang[1], Jinhyung Kim[1], Ahmadreza Ahmadi[1], Minkyu Choi[1] and Jun Tani[1,2,*]

[1]Korea Advanced Institute of Science and Technology, Daejeon, South Korea
[2]Okinawa Institute of Science Technology, Okinawa, Japan

[jungsik.hwang, kkjh0723, ar.ahmadi62, minkyu.choi8904, tani1216jp]@gmail.com

Abstract

This study presents a dynamic neural network model based on the predictive coding framework for perceiving and predicting the dynamic visuo-proprioceptive patterns. In our previous study [1], we have shown that the deep dynamic neural network model was able to coordinate visual perception and action generation in a seamless manner. In the current study, we extended the previous model under the predictive coding framework to endow the model with a capability of perceiving and predicting dynamic visuo-proprioceptive patterns as well as a capability of inferring intention behind the perceived visuomotor information through minimizing prediction error. A set of synthetic experiments were conducted in which a robot learned to imitate the gestures of another robot in a simulation environment. The experimental results showed that with given intention states, the model was able to mentally simulate the possible incoming dynamic visuo-proprioceptive patterns in a top-down process without the inputs from the external environment. Moreover, the results highlighted the role of minimizing prediction error in inferring underlying intention of the perceived visuo-proprioceptive patterns, supporting the predictive coding account of the mirror neuron systems. The results also revealed that minimizing prediction error in one modality induced the recall of the corresponding representation of another modality acquired during the consolidative learning of raw-level visuo-proprioceptive patterns.

*Index Terms*—Cognitive robotics, dynamic neural network, predictive coding, visuomotor learning.

I. INTRODUCTION

It is important to endow a robot with a capability of predicting and adapting to the dynamically changing environment [2, 3]. For instance, consider an imitation task between the two agents. This task not only requires perception and generation of actions but also requires higher-level cognitive skills, such as visuo-motor coordination and recognizing the demonstrator's intention by observing his/her gestures. By understanding the demonstrator's intention, the imitator can predict the demonstrator's behaviors

[*]Corresponding author

and also prepare their own behavior, resulting in successful interaction.

In this study, we introduce a dynamic neural network model called P-VMDNN (Predictive Visuo-Motor Deep Dynamic Neural Network) designed for processing and predicting raw-level dynamic visuo-proprioceptive patterns under the predictive coding framework [2-6]. The proposed model consists of two pathways (visual and proprioceptive pathway for perceiving and predicting the dynamic visual images and the perceptual outcome of the robot's intended actions respectively) and those two pathways are tightly coupled by means of the lateral connection at the highest layers in each pathway and end-to-end training of the dynamic visuo-proprioceptive patterns. The proposed model is an extension of our previous model [1, 7, 8] which was able to abstract and associate visual perception with proprioceptive information through a spatio-temporal hierarchical structure. In the current study, we extended the previous model under the predictive coding framework [2-5] to endow the model with several key features. First, the proposed model provides a mechanism for inferring an intention of the perceived visuo-proprioceptive patterns. Recognizing intention by observing other's behavior is one of the core skills required for social cognition [9-12]. One of the key assumptions in predictive coding is that the intention of the observed behavior can be inferred through minimizing the prediction error at the levels of a cortical hierarchy [5]. Similarly, the proposed model provides a mechanism for inferring underlying intention of the perceived visuo-proprioceptive patterns by minimizing prediction error through updating the internal states of the neurons at each level of the hierarchy in an online manner. Second, the proposed model is capable of proactively generating visuo-proprioceptive patterns with given intention states as well as mentally simulating the possible incoming dynamic visuo-proprioceptive sequences without the input from the external environment. Predicting the consequences of its own actions as well as others' through mental simulation is one of the important characteristics to interact in a dynamic environment [13, 14]. In addition to these features, the proposed model also inherits the key characteristics of the previous model [1, 7, 8], such as the hierarchical processing of raw-level dynamic visuo-proprioceptive patterns and tight coupling of perception and action.

A set of synthetic experiments were conducted to examine the proposed model and also to investigate possible biological mechanism of inferring higher-level intention of the perceived visuo-proprioceptive information. During the training process, the model was trained to predict both visual and proprioceptive outputs through consolidative learning of raw-level visuo-proprioceptive patterns acquired from the imitation task between the two agents. During the testing process, we examined the role of prediction error minimization in inferring underlying intention of the observed visuo-proprioceptive patterns as well as its effect on reconstructing the visuo-proprioceptive patterns through recalling of internal representations acquired during the training process. Particularly, we focused on how minimizing prediction error could elicit emergence of the mirror neuron systems as suggested in [5]. Furthermore, we also examined how the learned visuo-proprioceptive primitives can be reconstructed without the inputs from the external environment but by means of the top down process with given intention states (mental simulation). Although several previous studies [4, 15, 16] have introduced the computational model of the predictive coding framework, the current study, to our knowledge, is the first attempt to employ the predictive coding framework in the dynamic neural network model for processing and predicting raw-level dynamic visuo-proprioceptive patterns.

## II. DYNAMIC NEURAL NETWORK MODEL

*A. Overview*

The proposed model is called Predictive Visuo-Motor Deep Dynamic Neural Network (P-VMDNN) and it consists of two pathways (Fig. 1). The visual pathway perceives and predicts dynamic visual images and the proprioceptive pathway processes and predicts the perceptual outcome of the robot's intended actions. Note that the proposed model does not generate the actual actions but predicts the perceptual outcomes of the actions (i.e. proprioception). The robot's actual action can be generated with a low-level controller which controls the actuators with given predicted proprioception signals (desired joint angle values). Each pathway consists of a set of layers and the higher-level layers of each pathway are connected each other to link perception and action. The proposed model is an extension of our previous model [1, 7, 8] which was designed to abstract and associate visual perception with proprioceptive information through a spatio-temporally hierarchical structure. In the current study, we extended the previous model under the predictive coding framework [4, 5] to endow the model with several key features.

The proposed model can be characterized by the following aspects. First, the proposed model is a predictive model which can perceive and predict pixel-level images as well as the perceptual outcome of the robot's sequential actions. As a result, the proposed model provides a mechanism for minimizing the prediction error which is an essence of predictive coding [2, 4, 5]. In addition, as argued in [5], one of the important aspects of predictive coding is that the same structure is used to generate action and also to infer the cause of an action (i.e. intention) through minimizing prediction error. Similarly, the proposed model does not employ separate generative models, but the prediction error generated at the output layers back-propagates along the pathways in the model, utilizing the same hierarchical structure. Second, the proposed model is able to mentally simulate the possible incoming visuo-proprioceptive patterns without the inputs from the external environment. During the training, the model can learn the different intention states which can elicit proactive generation of the different dynamic visuo-proprioceptive sequences in a top-down process. By feeding its own visuo-proprioceptive output to the input of the model, the proposed model can generate dynamic visuo-proprioceptive patterns without the external information from the environment (i.e. mental simulation). Besides, the proposed model also inherits the essential characteristics of the previous model [1, 7, 8]. For example, the proposed model also utilizes the spatio-temporally hierarchical structure to process and to integrate raw-level visuo-proprioceptive sequences. Furthermore, the proposed model is also able to develop multimodal representation by which both perception and action are tightly intertwined within the system by means of the lateral connection at the higher-levels as well as the end-to-end training of the visuo-proprioceptive patterns.

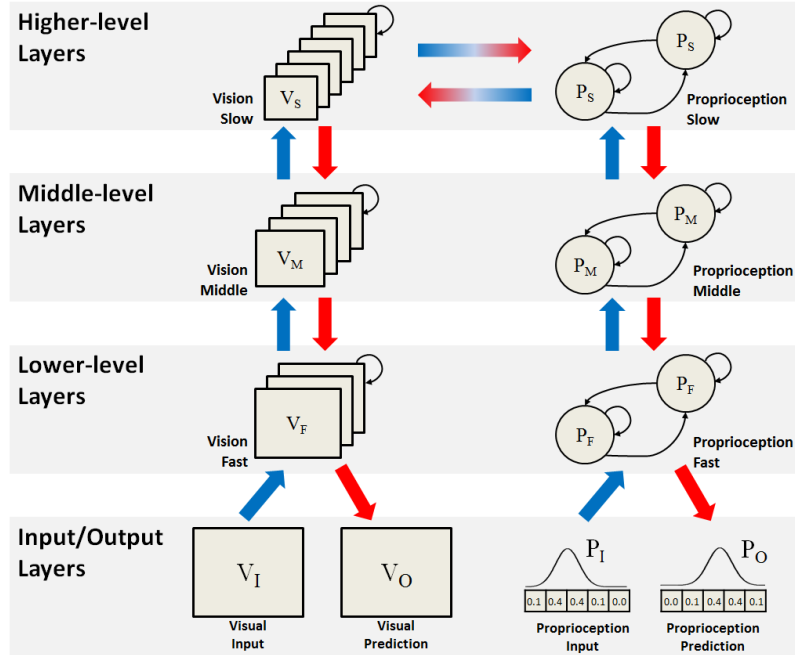

Fig. 1. The proposed model consists of two pathways: the visual pathway ($V_I$, $V_O$, $V_F$, $V_M$ and $V_S$) for perceiving and predicting dynamic visual image (left) and the proprioceptive pathway ($P_I$, $P_O$, $P_F$, $P_M$ and $P_S$) for processing and predicting robot's sequential action (right). The blue and red arrows indicate the bottom-up and the top-down pathways in the proposed model respectively. The arrows on the top represent the lateral connection between the highest layers in each pathway.

*B. Visual Pathway*

The visual pathway processes the dynamic visual patterns perceived by a robot and also predicts the visual image. In the proposed model, the variation of the P-MSTRNN (Predictive Multiple Spatio-Temporal Scales Recurrent Neural Network) model [17] was implemented to construct the visual pathway. The P-MSTRNN model is a dynamic neural network model based on the predictive coding framework and it is capable of perceiving and predicting the dynamic visual patterns. Each level in the P-MSTRNN model is imposed with different spatio-temporal constraints so that both spatial and temporal hierarchy can develop in the model. As a result, pixel-level dynamic images can be processed and generated through the hierarchical computation of the visual patterns. In our study, the original P-MSTRNN model was modified in a way that the feature maps and the context maps were merged into feature maps with recurrent loops.

The visual pathway in our model consists of five layers: Vision Input and Output ($V_I$, $V_O$), Vision Fast ($V_F$), Vision Middle ($V_M$) and Vision Slow ($V_S$). Each layer is composed of a group of feature maps retaining spatial information as well as temporal information of dynamic visual patterns. The Vision Input and Output layers consist of a single feature map containing the perceived and the predicted visual image respectively and they were connected to the Vision Fast layer. The feature maps in other layers in the visual pathway are connected asymmetrically to the feature maps in the neighboring layers and they were also equipped with the recurrent connection between the feature maps within the same layer. The layers in the visual pathway were imposed with different spatio-temporal constraints. To be more specific, the lower-level layers were assigned with smaller time constants and the shorter distance connectivity whereas the higher-level layers were assigned with bigger time constants and the longer distance connectivity. Consequently, progressively slower neural dynamics and longer distance connectivity (from the lower to the higher level) could be achieved as suggested in [17].

The internal states $u_i^t$ and the dynamic activation $v_i^t$ of the neurons in the visual pathway at the time step $t$ are computed using the following equations:

$$u_i^t = \left(1 - \frac{1}{\tau_i}\right)u_i^{(t-1)} + \begin{cases} \frac{1}{\tau_i}\left(\sum_{j \in V_M \vee V_S} k_{ij} * v_j^{t-1} + \sum_{k \in P_S} k_{ik} * y_k^{t-1} + b_i\right) & if\ i \in V_S \\ \frac{1}{\tau_i}\left(\sum_{j \in V_F \vee V_M \vee V_S} k_{ij} * v_j^{t-1} + b_i\right) & if\ i \in V_M \\ \frac{1}{\tau_i}\left(\sum_{j \in V_F \vee V_M} k_{ij} * v_j^{t-1} + \sum_{k \in V_I} k_{ik} * V_k^t + b_i\right) & if\ i \in V_F \\ \frac{1}{\tau_i}\left(\sum_{j \in V_F} k_{ij} * v_j^t + b_i\right) & if\ i \in V_O \end{cases} \quad (1)$$

$$v_i^t = \begin{cases} \tanh(u_i^t) & if\ i \in V_O \\ 1.7159 \times \tanh\left(\frac{2}{3}u_i^t\right) & otherwise, \end{cases} \quad (2)$$

Where $\tau$ is the time constant, * is the convolution operator, $k_{ij}$ is the kernel connecting $j$th feature map in $V_j$ with the $i$th feature map in the current layer, b is the bias and $V^t$ is an input visual image. Note that the transposed convolution was performed for the top-down and the lateral connections. To enhance the model's performance, we used the hyperbolic tangent recommended in [18] for the activation function of the context neurons whereas a typical hyperbolic tangent was used for the activation function of the outputs and the initial states.

*C. Proprioceptive pathway*

The proprioceptive pathway processes and predicts the perceptual outcomes of the robot's sequential actions and it was implemented using a dynamic neural network called Multiple Timescales Recurrent Neural Network (MTRNN) [19]. The MTRNN model is a type of continuous time recurrent neural network with leaky integrator neurons and it can be characterized by a self-organizing functional hierarchy achieved by the different time scales dynamics [19]. The MTRNN model supports the hierarchical representation of actions [5, 19] in a way that the robot's sequential action can be modelled into a set of primitives and those primitives can be combined flexibly to form another actions [19, 20].

In our model, the proprioceptive pathway consists of five layers: Proprioception Input and Output ($P_I$, $P_O$), Proprioception Fast ($P_F$), Proprioception Middle ($P_M$) and Proprioception Slow ($P_S$). The Proprioception Input and Proprioception Output layers consist of the softmax neurons representing the encoder values of the robot's joints and they are connected to the Proprioception Fast layer. The neurons in other layers in the proprioceptive pathway were connected asymmetrically to the neurons in the neighboring layers and they were also equipped with the recurrent connection between the neurons within the same layer. Different temporal constraints were imposed on each layer in the proprioceptive pathway. More specifically, the progressively larger time constants were assigned from the lower levels to the higher levels in the proprioceptive pathway. Consequently, the neurons in the lower layers with a smaller time constant exhibit the faster dynamics compared to the ones in the higher layers assigned with the bigger time constants. This progressively slower neural dynamics were found to be important to form functional hierarchy in the previous studies [1, 7, 19].

The internal states $u_i^t$ and the dynamic activation $y_i^t$ of the neuron in the proprioceptive pathway at the time step $t$ are computed by the following equations:

$$u_i^t = \left(1 - \frac{1}{\tau_i}\right) u_i^{t-1} + \begin{cases} \frac{1}{\tau_i}\left(\sum_{j \in P_M \vee P_S} w_{ij} y_j^{t-1} + \sum_{k \in V_S} k_{ik} * v_k^t + b_i\right) & \text{if } i \in P_S \\ \frac{1}{\tau_i}\left(\sum_{j \in P_F \vee P_M \vee P_S} w_{ij} y_j^{t-1} + b_i\right) & \text{if } i \in P_M \\ \frac{1}{\tau_i}\left(\sum_{j \in P_F \vee P_M} w_{ij} y_j^{t-1} + \sum_{k \in P_I} w_{ik} P_k^t + b_i\right) & \text{if } i \in P_F \\ \frac{1}{\tau_i}\left(\sum_{j \in P_F} w_{ij} y_j^t + b_i\right) & \text{if } i \in P_O \end{cases} \quad (3)$$

$$y_i^t = \begin{cases} \frac{\exp(u_i^t)}{\sum_{j \in P_O} \exp(u_j^t)} & \text{if } i \in P_O \\ 1.7159 \times \tanh\left(\frac{2}{3} u_i^t\right) & \text{otherwise,} \end{cases} \quad (4)$$

Where $w_{ij}$ is the weight from the $j$th neuron to the $i$th neuron, $k$ is the kernel connecting $k$th feature map in $V_S$ with the $i$th neuron in $P_S$ and $P^t$ is a proprioception input represented in the softmax form.

*D. Generation Mode*

At each time step $t$, a pixel-level grayscale image $V^t$ obtained from the robot's camera was given to the vision input layer ($V_I$) and the encoder values of the robot's joints in the softmax representation $P^t$ were given to the proprioception input layer ($P_I$). Then, the internal states and the dynamic activation of the neurons at each layer were computed using the Eq. (1) ~ (4).

Two types of the generating visuo-proprioceptive patterns were used in our study. The first method is called an open-loop generation or the sensory entrainment. In this method, the visuo-proprioceptive input to each pathway was given from the external sources such as cameras and encoders, entraining the neural dynamics of the model. Another method is called a closed-loop generation in which the visuo-proprioceptive prediction generated at the current time step is fed back to the model as an input in the next time step. Consequently, this method does not require the inputs from the external environment, enabling the mental simulation of possible incoming visuo-proprioceptive sequences [21]. During the training process, only the closed-loop generation was used whereas both open-loop and closed-loop generation were compared in the testing cases.

*E. Training Mode*

The objective of the training was to optimize the model's learnable parameters to minimize the error at the output layer of each pathway. The model was trained in a supervised end-to-end learning in which the raw-level visuo-proprioceptive patterns obtained from the tutoring process were given to the network. In the tutoring process prior to the training, visual images obtained from the robot's camera were jointly collected with the values of the encoders in the robot's joints at each time step while the robot was manually operated by the experimenter. Backpropagation through time (BPTT) [22] was employed to obtain the values of the learnable parameters including kernels, weights, biases and the initial states of the neurons in each layer except the input and output layers ($V_I$, $V_O$, $P_I$ and $P_O$). By having the different initial states for each exemplar, it was possible to endow the model with the proactive visuo-proprioceptive pattern generation and the mental simulation capabilities. In this study, the model was trained to generate one-step look ahead visuo-proprioceptive predictions and the error during the training mode was defined as the sum of the errors in the visual pathway ($E_V$) and the proprioceptive pathway ($E_P$) as follow:

$$E = E_V + E_P \qquad (5)$$

$$E_V = \sum_t \sum_i (\bar{v}_i^t - v_i^t)^2 \qquad (6)$$

$$E_P = \sum_t \sum_i \bar{y}_i^t \log \frac{\bar{y}_i^t}{y_i^t} \qquad (7)$$

Where $\bar{v}$ and $\bar{y}$ are the values of the teaching signal for visual and proprioceptive prediction respectively. The model was trained in the closed-loop manner in the training model. That is, the prediction for vision and proprioception generated at the current time step were used as the feedback input in the next time step.

*F. Minimizing Prediction Error through Inferring Internal States*

One of the key capabilities of the proposed model is that it provides the mechanism for minimizing prediction error in an online manner. Minimizing prediction error is an essence of the predictive coding and it can account for the mirror neuron system [23, 24] in a way that the cause of an observed action (such as intention and goal) can be inferred by minimizing the prediction error during action observation [5]. Similarly, the proposed model provides the mechanism called an error regression scheme (ERS) [21, 25] in which the higher-level intention can be inferred through minimizing the prediction errors in an on-line manner.

In the ERS, the visuo-proprioceptive predictions are generated in a closed-loop manner with given intention represented by the internal states (top-down process). Then, the prediction error generated at the output layers back-propagates along the pathways to update the internal states of the neurons in the direction of minimizing the prediction errors at the output layers (bottom up process) [21, 25]. This top-down and bottom-up processes iterate during the ERS to minimize the prediction error and ultimately, to infer the underlying intention of an observed visuo-proprioceptive patterns. To be more specific, at each time step $t$, the prediction error defined by the discrepancy between the predicted and perceived visuo-proprioceptive patterns back-propagates through the temporal window indicating the immediate past (from the current time step $t$ to $t$-$W$ where W is the size of the window). The prediction errors generated within the temporal window are used to update the initial state of the window $U_{t-w}$ in the direction of minimizing prediction error generated within the window, resulting in changes to dynamic activation of the neurons of all level as well as visuo-proprioceptive predictions inside the window. By means of minimizing prediction error generated in the immediate past, the model is able to update the current intention to match the one behind the perceived visuo-proprioceptive patterns. Note that only the initial values of the window are updated during the ERS and the model's learnable parameters such as weights, kernels and biases are initialized with the ones obtained from the training and remained fixed during the ERS. For the detailed description about the ERS, please see [21, 25].

III. METHODS

*A. Experiment Settings*

The iCub simulator [26] was used in our study to obtain the visuo-proprioceptive patterns. iCub [27] is a child-like humanoid robot designed for cognitive robotics research and the iCub simulator [26] provides the flexible interface for accessing robot's devices such as cameras, actuators and other sensors in the simulation environment, making it a suitable platform for studying cognitive and developmental robotics [1, 7, 16, 28].

In our study, the robot was trained to imitate the hand-waving gestures of another robot displayed on the screen in the simulation environment (Fig. 2 (a)). Note that the robot predicted not only its own movement (prediction of own proprioception) but also the movement of another robot on the screen (visual prediction). A total number of 16 visuo-proprioceptive patterns were used in the training. The patterns consisted of a set of visual images showing the gestures of another robot on the screen as well as corresponding joint angles recorded during the tutoring process in which the robot was manually operated to imitate the hand-waving gestures of another robot on the screen. The hand-waving gestures in the imitation task were comprised of three different types of gestures categorized in terms of the arm that moved first at the onset of the task: left, right and both arms. For each type of the gesture, the amplitude of the waving was systematically controlled for each arm, such as full, half and no hand-waving. Regarding the visual images in the training data, another robot's gesture displayed on the screen was perceived through the camera embedded in the robot's left eye and the obtained images were converted to grayscale, resized to 64 (w) × 48(h) and normalized to −1 to 1. To generate the robot's behavior, we used the two joints (left and the right elbow) and then, those joint values were converted to the two groups of softmax representation consisting of 10 softmax neurons each.

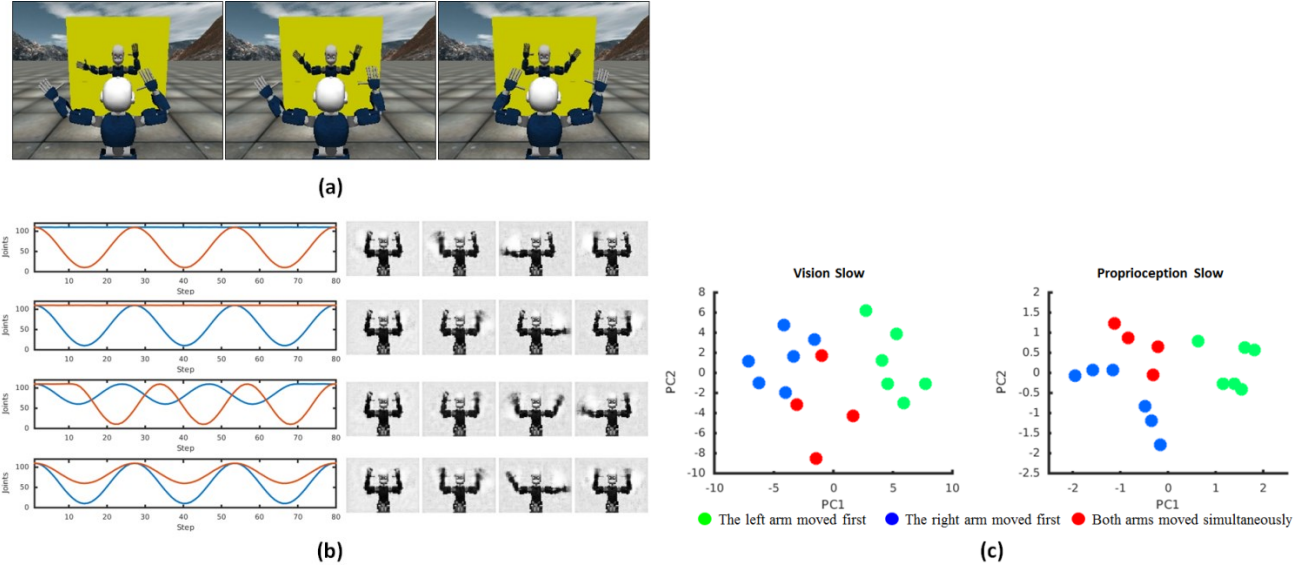

Fig. 2. (a) The experiment setting for collecting the visuo-proprioceptive data. The robot was tutored to imitate the gesture of another robot displayed on the screen (b) Examples of the visuo-proprioceptive patterns in the closed-loop generation (mental simulation) in Experiment 1. The left plots show the two joints' angle values and the right figures show the visual predictions. (c) PCA plot on the initial states of each primitive in the highest layers ($V_S$ and $P_S$). The horizontal and the vertical axis indicate the first and the second principal component respectively and the colors denote the characteristics of the visuo-proprioceptive patterns (i.e. the arms that moved first at the onset).

*B. Network Settings*

The input and output layers of the visual pathway ($V_I$ and $V_O$) consisted of a single feature map (64 (w) × 48 (h)) containing the perceived and the predicted visual image respectively. The input and output layer of the proprioceptive pathway ($P_I$ and $P_O$) consisted of 20 neurons in the softmax form indicating the two analog values (i.e. joint angle values) represented in the 10 softmax units each. The input and output layers in each pathway were assigned with the lowest time constants ($\tau = 1$) whereas the highest levels in each pathway were assigned with the highest time constants ($\tau = 8$). Table 1 shows the network's parameter settings used in our experiments including the time constants, the size of the feature maps, kernels and weights. The values for those parameters were found empirically. At the onset of the training, the learnable parameters of the network were initialized to the neutral values. We trained the model for 40,000 epochs using the ADAM optimizer [29] with the learning rate of 0.001 in Tensorflow [30].

Table 1. The network settings in our experiment

| | Layer | Time Constants | Feature Maps | | Top-Down Kernel | | Bottom-Up Kernel | | Recurrent Kernel | | Lateral Kernel | |
| --- | --- | --- | --- | --- | --- | --- | --- | --- | --- | --- | --- | --- |
| | | | Number | Size | Size | Stride | Size | Stride | Size | Stride | Size | Stride |
| Visual Pathway | $V_F$ | 2 | 4 | 60×44 | 4×4 | 2,2 | 5×5 | 1,1 | 2×2 | 1,1 | - | - |
| | $V_M$ | 4 | 8 | 29×21 | 5×5 | 2,2 | 4×4 | 2,2 | 2×2 | 1,1 | - | - |
| | $V_S$ | 8 | 12 | 13×9 | - | - | 5×5 | 2,2 | 2×2 | 1,1 | 13×9 | 1,1 |
| | Layer | Time Constants | Number of Neurons | | Top-Down Weights | | Bottom-Up Weights | | Recurrent Weights | | Lateral Kernel | |
| Proprioceptive pathway | $P_F$ | 2 | 30 | | 30×20 | | 30×20 | | 30×30 | | - | - |
| | $P_M$ | 4 | 20 | | 20×10 | | 20×30 | | 20×20 | | - | - |
| | $P_S$ | 8 | 10 | | - | | 10×20 | | 10×20 | | 13×9 | 1,1 |

## IV. RESULTS

We conducted two experiments to examine the key characteristics of the proposed model. In the first experiment, we evaluated the model's capability of proactively generating and mentally simulating visuo-proprioceptive patterns. In the second experiment, the proposed model's capability of inferring intention through minimizing prediction error in an online manner was evaluated.

### A. Experiment 1. Top-Down Proactive Generation of the Visuo-Proprioceptive Patterns

In the first experiment, we examined the model's capability of proactively generating and mentally simulating the possible incoming visuo-proprioceptive sequences with a given intention state. At the onset of the generation mode, the initial states of the neurons in the entire layers were initialized to the ones obtained after the training for each primitive. Then, the same initial visuo-proprioceptive values representing the home position were fed to the input layers in each pathway for every primitive. Then, the model generated the dynamic visuo-proprioceptive patterns for each primitive in a closed-loop manner afterwards.

The result showed that the model was able to generate the dynamic visuo-proprioceptive patterns successfully without the inputs from the external sources, illustrating the mental simulation capability of the proposed model. Fig. 2 (b) shows some of the examples of the generated visuo-proprioceptive patterns (see the supplementary video at https://youtu.be/Sf_AuO2yFws). As can be seen from the figure, the visual and proprioceptive predictions were coherent in the whole primitives and this result implies that perception and action were tightly coupled in the proposed model. We also conducted principal component analysis (PCA) on the initial states of the higher-level layers ($V_S$ and $P_S$) in each pathway. The X axis and the Y axis in Fig. 2 (c) indicate the first and the second principal components respectively and the colors indicate the characteristics of the gesture. The figure illustrates that the higher-level 'intention' was self-organized in a way that it reflected the characteristics of the primitives. In sum, the first experiment illustrated one of the essences of the proposed model that is with a given prior intention, the model was able to mentally simulate possible incoming visuo-proprioceptive patterns in synchrony.

### B. Experiment 2. Minimizing Prediction Error through Inferring Internal States

In Experiment 2, we examined another essence of the proposed model – inferring the internal state by means of minimizing the prediction error since the proposed model generated the prediction in the two pathways, we conducted the experiment under the two different conditions: minimizing visual prediction error (PE) and minimizing proprioceptive PE. In the first condition (visual PE minimization), we assumed that the robot observed a gesture of another robot and the prediction error between the perceived and the predicted gesture of another robot was minimized through updating the internal states. In the second condition (proprioceptive PE minimization), we assumed that the desired kinematics were provided from the external source (e.g., the experimenter) and the discrepancy between the predicted and the actual proprioceptive outputs was minimized through updating the internal states. This condition emulates the case in which the robot's arms are manually operated by the experimenter and the robot is imagining the visual inputs that correspond to the given proprioception teaching signals. In both conditions, both visual and proprioceptive outputs were generated in the closed-loop manner, meaning that the predicted visuo-proprioceptive pattern was used as the input in the next time step. In addition, we also evaluated the model's performance when the PE minimization was not employed (i.e. sensory entrainment) in each condition. In the sensory entrainment in the first condition, the visual input from the external source (i.e. camera) was given to the input of the visual pathway at each time step and the proprioceptive output was generated in the closed-loop manner. In the sensory entrainment in the second condition, the proprioception input from the external source (i.e. encoder) was given to the input of the proprioceptive pathway at each time step and the visual output was generated in the closed-loop manner.

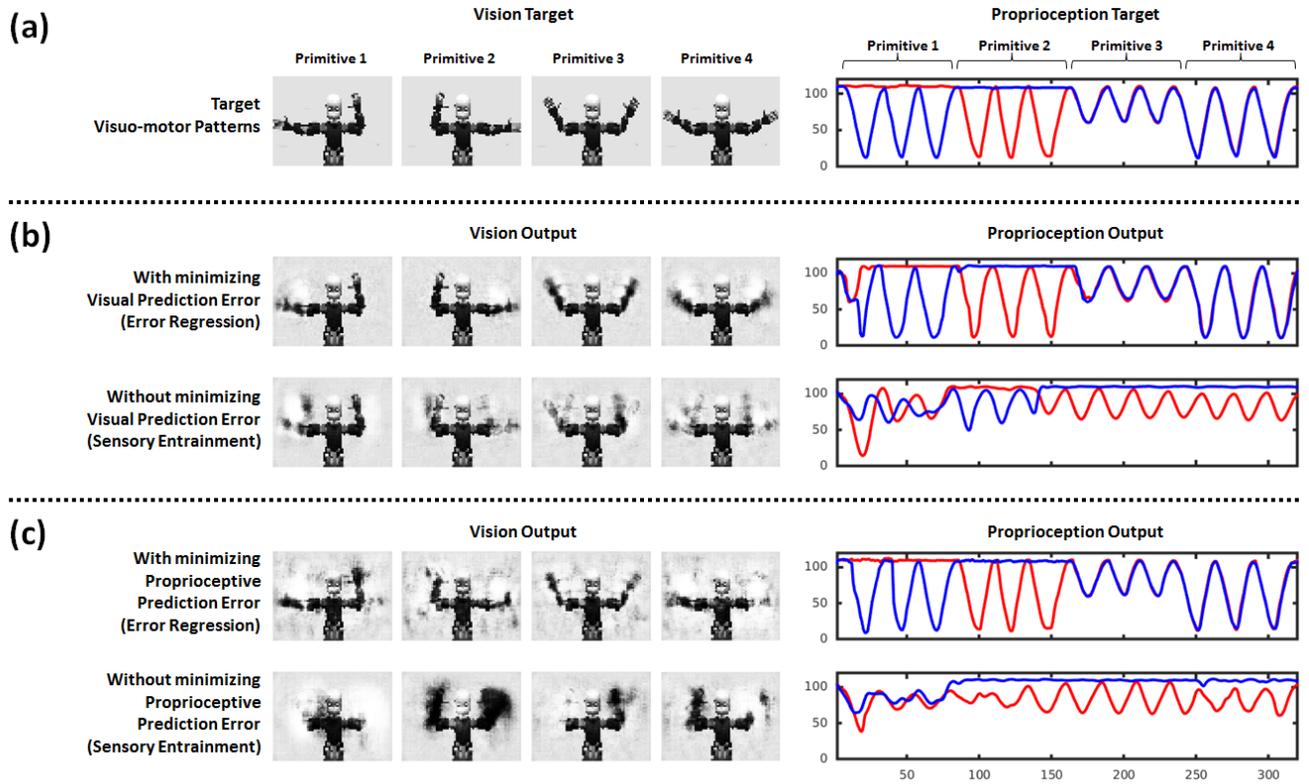

Fig. 3. (a) The visuo-proprioceptive patterns presented in Experiment 2. (b) The visuo-proprioceptive output in the first condition in Experiment 2 (Visual Prediction Error Minimization) (c) The visuo-proprioceptive output in the second condition in Experiment 2 (Proprioceptive Prediction Error Minimization). For each condition, the visual output (left) and the proprioception output (right) are depicted. The numbers at the bottom in the right figures indicate the time step from the beginning to the end of the task, and the colors denote the joint angle values of the robot.

In the testing, a dynamic visuo-proprioceptive sequence consisting of the four concatenated training patterns but with certain fluctuation in the proprioceptive signal was presented to the model. The kernels, weights and biases were initialized to the values that obtained from the training and the internal states were initialized to the neutral values at the onset of the testing. The size of the temporal window was chosen as 30 and the initial states of the window were updated 50 times for each time step using the ADAM optimizer [29] with the learning rate of 0.1. As described in Section II. F, only the initial states of the temporal window were updated in the direction of minimizing the prediction error while the other learnable parameters were remained fixed.

The results clearly showed the importance of prediction error minimization. In the first condition (visual PE minimization), the model was able to predict the gesture of another robot through updating of the initial states while observing the gestures (Fig. 3 (b)). In addition, the model was also able to imitate another robot's gesture by predicting its own proprioceptive outputs that correspond to the visual prediction. In contrast, the model was not able to predict neither the gesture nor the proprioceptive outputs in the sensory entrainment case. This result indicates the importance of prediction error minimization in social interaction between two agents. The similar results were also found in the second condition (minimizing proprioceptive PE). As can be seen from Fig. 3 (c), the model successfully minimized the prediction error in the proprioceptive pathway. More interestingly, the model was also able to imagine the visual inputs that correspond to the current proprioceptive prediction. In contrast, the model was not able to predict visuo-proprioceptive patterns in the sensory entrainment case, highlighting the importance of prediction error minimization (See the supplementary video).

In order to clarify the internal dynamics of the model, we conducted PCA on the dynamic activation of the neurons at the higher-level of the visual pathway ($V_S$) and the higher and the lower level of the proprioceptive pathway ($P_S$ and $P_F$) in the visual PE minimization condition. In Fig. 4, the internal representations emerged after the training and ones emerged during the testing phases are depicted. The X axis and the Y axis indicate the first and the second principal components respectively. As can be seen from the figure, the internal representation emerged through optimizing visual prediction error was similar to the ones in the training phase, suggesting that the underlying intention of the observed action was successfully recognized. At the earliest phase of the ERS, the internal representations of the primitive 1 in the ERS showed a slight difference compared to the ones in the training phase. This might be due to the smaller size of the temporal window at the beginning of the ERS where the current time step t is smaller than the size of the window (W). However, the internal representations in the ERS converged to the ones during the training process afterwards. In other words, the model was able to update its own intention to imitate demonstrator's behavior through minimizing the error between the predicted and the actual movements of the demonstrator. This finding supports the role of predictive coding in accounting for the mirror neuron systems as discussed in [5]. In addition, the internal representations emerged in the proprioceptive pathway during the ERS were similar to the ones emerged during the training. It is assumed that minimizing visual prediction error induced the recall of the corresponding proprioceptive representations that were acquired during the training, resulting in the proper mental simulation of the proprioceptive outputs. This finding is similar to the one in [16] that showed the missing sensorimotor signals can be retrieved. Moreover, this finding is in line with the previous studies in neuroscience that showed the perceiving actions that exist in the motor repertoire modulated the activation of cortical motor regions [31, 32]. In sum, the findings in Experiment 2 illustrated the importance of prediction error minimization in inferring intention as well as recalling the visuo-proprioceptive representations that were acquired in the training process.

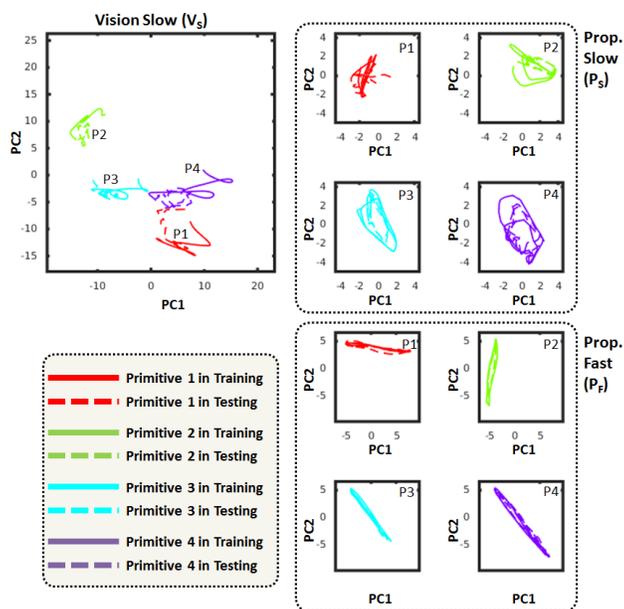

Fig. 4. The internal representations of the primitives emerged during training (solid lines) and the testing (dashed lines) in the first condition (Visual Prediction Error Minimization) in Experiment 2. The horizontal and the vertical axis indicate the first and the second principal component respectively. The colors denote the primitives used in the experiment.

## V. Conclusion

In this study, we introduced the dynamic neural network model called P-VMDNN (Predictive Visuo-Motor Deep Dynamic Neural Network) for visuomotor learning based on the predictive coding framework. The experimental results illustrated several core characteristics of the proposed model. First, the model was able to generate visuo-proprioceptive patterns with given intention through top-down mechanism, resulting in mental simulation of possible incoming dynamic visuo-proprioceptive sequences without external inputs. Second, the results also revealed that minimizing prediction error played an important role in inferring intention as well as recalling visuo-proprioceptive representations acquired during the training. Higher-level intention could be inferred through updating the internal states in the direction of minimizing the prediction error, illustrating the mirror neuron-like mechanism of the proposed model. By minimizing the prediction error in one modality, the model was able to imagine the perceptual sequences of another modality through updating the internal states that induced the recall of the corresponding representation of another modality. In our future study, the model will be examined under more complex tasks including learning of compositional visuo-proprioceptive patterns for an in-depth understanding of visuomotor learning under the predictive coding framework.

## Acknowledgment

This work was supported by the ICT R&D program of MSIP/IITP. [2016(R7117-16-0163), Intelligent Processor Architectures and Application Softwares for CNN-RNN]